\documentclass[letterpaper, 10 pt, conference]{ieeeconf}  
\usepackage{bm}
\IEEEoverridecommandlockouts

\usepackage{graphics} 
\usepackage{epsfig} 
\usepackage{mathptmx} 
\usepackage{times} 
\usepackage{amsmath} 
\usepackage{amssymb}  
\usepackage{caption}
\usepackage{subcaption}
\usepackage{xcolor}
\usepackage{cite}
\usepackage{float}
\usepackage{soul}
\usepackage{multirow}

\usepackage{tikz,xcolor,hyperref}
\definecolor{lime}{HTML}{A6CE39}
\DeclareRobustCommand{\orcidicon}{
	\begin{tikzpicture}
	\draw[lime, fill=lime] (0,0) 
	circle [radius=0.16] 
	node[white] {{\fontfamily{qag}\selectfont \tiny ID}};
	\draw[white, fill=white] (-0.0625,0.095) 
	circle [radius=0.007];
	\end{tikzpicture}
	\hspace{-2mm}
}
\foreach \x in {A, ..., Z}{\expandafter\xdef\csname orcid\x\endcsname{\noexpand\href{https://orcid.org/\csname orcidauthor\x\endcsname}
			{\noexpand\orcidicon}}
}

\title{\LARGE \bf{On Improving Multimodal Pedestrian Trajectory Prediction with CVAE: A Study on Benchmark and Robot Data}}

\author{Yuzhou Liu\orcidF \emph{Student Member, IEEE},
and Cristina Olaverri-Monreal\orcidE{} \emph{Senior Member, IEEE}%
\thanks{\textsuperscript{1}Dept. Intelligent Transport Systems, Johannes Kepler University Linz, Altenberger Straße~69, 4040~Linz, Austria. 
	\texttt{\{yuzhou.liu, cristina.olaverri-monreal\}@jku.at}}%
}

\begin{document}

\maketitle
\thispagestyle{empty}
\pagestyle{empty}

\begin{abstract}
Accurate pedestrian trajectory prediction is crucial for autonomous systems operating in complex environments, such as modular buses and delivery robots in suburban or semi-structured areas. Social Spatio-Temporal Graph Convolutional Neural Networks (Social-STGCNN) have shown strong performance by modeling social interactions; however, producing diverse and well-calibrated future trajectories remains challenging.
In this work, we build on a Social-STGCNN backbone and introduce a Conditional Variational Autoencoder (CVAE)-based probabilistic formulation to explicitly model multimodal future trajectories. We evaluate the method on the ETH and UCY pedestrian trajectory datasets as well as on a real-world pedestrian dataset collected by a mobile robot. Results show moderate gains on public benchmarks, but more consistent endpoint accuracy and improved trajectory diversity across different crowd configurations. Evaluation on robot-collected data further demonstrates the approach’s effectiveness beyond curated benchmarks and supports its applicability in practical deployments.
\end{abstract}

\section{Introduction}\label{sec:1}

In recent years, advancements in Automated Vehicle (AV) technology have increased the demand for robust perception systems and enhanced pedestrian safety measures. Anticipating pedestrian movement is crucial for adaptive urban mobility, particularly in applications such as modular buses and delivery robots operating in pedestrian-rich environments. Accurate trajectory prediction enables these automated systems to make informed decisions, plan safe routes, and navigate complex urban environments with improved reliability.
Pedestrian trajectory prediction remains challenging due to the inherent variability and unpredictability of human behavior. Unlike vehicles, pedestrians exhibit diverse motion patterns influenced by personal intentions, social interactions, and cultural norms \cite{schwarz2015safety}. Therefore, effective prediction models must be capable of handling uncertainty and adapting to dynamic and evolving scenarios.

Many researchers \cite{alahi2016social,gupta2018social,sadeghian2019sophie,li2019conditional,zhang2019sr, kosaraju2019social} have studied pedestrian behavior in open environments such as sidewalks, residential areas, and campuses. These settings are also relevant for systems like delivery robots and modular buses, where onboard perception observes pedestrians in close proximity and under varying crowd densities.
To model such interactions, deep learning (DL) approaches typically rely on observing temporal sequences of pedestrian motion. However, many existing methods employ complex architectures such as Recurrent Neural Networks (RNNs) and Transformers, which can be computationally expensive for real-time deployment.

Social-STGCNN \cite{mohamed2020social} addresses this by modeling pedestrians as spatio-temporal graphs and applying convolutional architectures to capture both spatial and temporal dependencies. Building upon the ST-GCNN framework \cite{yan2018spatial}, it extracts graph-based features efficiently and predicts future trajectories using a temporal convolutional decoder. This approach achieves strong performance while maintaining computational efficiency. Several subsequent works \cite{zhou2021ast, cao2021spectral, xu2022adaptive, li2022graph, pang2022bayesian} further build upon this idea.

Despite these advances, modeling the uncertainty and multimodality of future pedestrian trajectories remains challenging. In many real-world scenarios, multiple plausible future paths may exist, especially in crowded or semi-structured environments.
To address this, we introduce a probabilistic formulation based on a CVAE. 
Our approach builds upon a Social-STGCNN backbone and incorporates a CVAE framework to model the distribution of possible future trajectories. 
We selected Social-STGCNN due to its computational efficiency and convolutional graph formulation, making it suitable for real-time robotic systems and a practical backbone for studying probabilistic extensions.
Inspired by the CVAE-based human motion prediction model HuMoR \cite{rempe2021humor}, we leverage both past and future trajectories during training to learn a latent representation, while relying only on past observations during inference.
%
While several works have explored CVAE-based approaches for trajectory prediction \cite{choi2021drogon, yao2021bitrap, zhou2022dynamic, gao2022social, yue2022human, xu2023social, yang2023sgamte, zhu2023tri}, many of them either focus on predicting individual trajectories or adopt complex architectures, which highly increase the resources needed. In contrast, our work focuses on integrating a lightweight probabilistic module into a graph-based model, allowing joint prediction of multiple pedestrians while maintaining efficiency.
Rather than proposing a fundamentally new architecture, this work studies how conditional variational inference can be effectively integrated into a graph-based trajectory predictor and evaluates its behavior on both standard benchmarks and real-world data. 

We evaluate the proposed method on the ETH and UCY datasets, as well as on a pedestrian dataset collected by a mobile robotic platform. While improvements on public benchmarks are moderate, the model demonstrates more consistent endpoint accuracy and improved trajectory diversity across different crowd configurations. Results on robot-collected data further support the applicability of the approach in real-world scenarios relevant to modular bus and robotic systems.

Overall, this paper makes the following contributions:

\begin{itemize}
    \item We present a practical recipe for integrating CVAE into variable-size multi-agent graphs without pooling and flattening, preserving permutation and size flexibility.
    \item We demonstrate a lightweight probabilistic upgrade of Social-STGCNN that improves endpoint robustness with minimal latency increase, keeping real-time feasibility.
    \item We provide an empirical evaluation on both standard benchmarks and a real-world robot-collected dataset, highlighting the practical applicability of the approach.
\end{itemize}

This paper is structured as follows: In Section \ref{sec:2} we introduce Related Work in the field of pedestrian trajectory prediction along with relevant DL architectures.
In Section \ref{sec:3} we describe the proposed architecture followed by Experimental Setup in Section \ref{sec:4}. 
Section \ref{sec:5} presents the results and Section \ref{sec:6} concludes the paper outlining future research.

\section{Related Work}\label{sec:2}

\textbf{Social concepts based DL models}
Social-LSTM \cite{alahi2016social} was one of the earliest deep learning approaches for pedestrian trajectory prediction that explicitly modeled social interactions. It employed a Recurrent Neural Network (RNN) to capture individual motion patterns and introduced a social pooling mechanism to aggregate interactions between pedestrians, assuming a bivariate Gaussian distribution over future trajectories. However, its reliance on sequential modeling leads to limited computational efficiency.
Social-GAN \cite{gupta2018social} extended this idea by introducing a Generative Adversarial Network (GAN) to generate more diverse trajectories. While this improves multimodality, GAN-based methods are known to suffer from training instability and mode collapse, which can limit their robustness. 
SoPhie \cite{sadeghian2019sophie} further incorporated visual scene context and attention mechanisms, but at the cost of increased model complexity. 
Similarly, CGNS \cite{li2019conditional} replaced LSTMs with GRUs to improve efficiency, while SR-LSTM \cite{zhang2019sr} and Social-BiGAT \cite{kosaraju2019social} introduced weighting and graph-based interaction modeling to better capture social influence. Despite these improvements, most of these approaches rely on sequential architectures, which can be computationally demanding and less suitable for real-time applications.

\textbf{Social-STGCNN extensions}
With the development of Graph Convolutional Networks (GCNs) \cite{kipf2016semi}, spatio-temporal graph-based models \cite{yu2017spatio,yan2018spatial} have become a popular framework for trajectory prediction. Social-STGCNN \cite{mohamed2020social} represents pedestrians as nodes in a spatio-temporal graph and applies convolutional operations to jointly model spatial interactions and temporal dynamics. By replacing recurrent structures with convolutional ones, it achieves both strong performance and improved efficiency.
Several works have extended this framework. AST-GNN \cite{zhou2021ast} incorporates attention mechanisms to enhance feature aggregation, while SpecTGNN \cite{cao2021spectral} explores spectral representations of spatio-temporal graphs. Transferable GNN \cite{xu2022adaptive} focuses on domain adaptation to improve generalization across environments, and \cite{li2022graph} introduces multi-scale graph-based transformers with trajectory smoothing.
While these methods improve feature extraction and representation within the graph-based paradigm, 
they generally follow a deterministic or unimodal prediction pipeline, where a fixed set of observations is mapped to a single predicted trajectory or a limited parametric distribution.
%
As a result, modeling the inherent uncertainty and multimodality of future pedestrian motion remains challenging, particularly in crowded or ambiguous scenarios.

\textbf{CVAE based works}
Conditional Variational Autoencoders (CVAEs) have been widely explored for modeling uncertainty in trajectory prediction. DROGON \cite{choi2021drogon} and BiTraP \cite{yao2021bitrap} introduce goal-conditioned frameworks that first estimate future endpoints and then generate trajectories. However, their performance depends heavily on accurate goal prediction. 
Social-DualCVAE \cite{gao2022social} improves interaction modeling through dual recognition and prior networks, while NSP-SFM \cite{yue2022human} incorporates goal sampling and environmental features, both at the cost of increased model complexity.
Other approaches focus on enhancing multimodality through generative modeling. CVAE-GAN \cite{zhou2022dynamic} combines CVAE with adversarial training, introducing additional training challenges. Social-CVAE \cite{xu2023social} focuses on single-pedestrian prediction, limiting its applicability in multi-agent settings. SGAMTE-Net \cite{yang2023sgamte} and Tri-HGNN \cite{zhu2023tri} introduce more complex architectures to improve diversity and interaction modeling, but often require additional design choices such as predefined interaction structures or multiple coupled networks.

In contrast to these approaches, our work focuses on a lightweight integration of a CVAE framework into a graph-based predictor. This allows joint modeling of multiple pedestrians while maintaining computational efficiency, and provides a practical way to capture multimodal future trajectories within a convolutional spatio-temporal architecture.

\section{Methodology}\label{sec:3}

In this section, we describe our model for pedestrian trajectory prediction. 
Following previous works \cite{rempe2021humor,mohamed2020social}, we combine a Social-STGCNN backbone with a CVAE framework. 
We follow the graph construction and spatio-temporal representation introduced in Social-STGCNN \cite{mohamed2020social}, where pedestrian trajectories are represented as spatio-temporal graphs with fully connected interactions at each time step.

\begin{figure*}[t]
    \centering
    \includegraphics[width=0.95\textwidth]{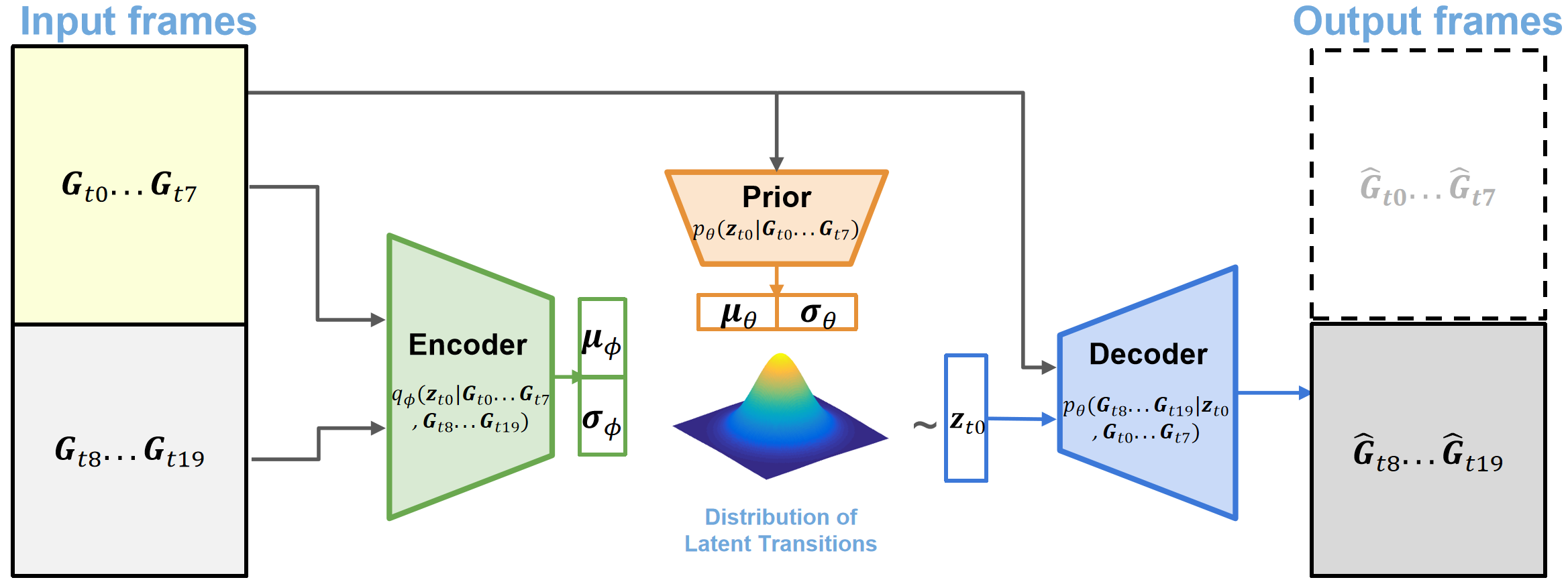}
    \caption{This figure illustrates our CVAE architecture. During training, given ground truth of frame of $\mathbf{G}_{t0}...\mathbf{G}_{t19}$, the model reconstructs $\mathbf{\hat{G}}_{t0}...\mathbf{\hat{G}}_{t19}$ by sampling from the encoder distribution. At test time we can generate whole 20 frames from $\mathbf{G}_{t0}...\mathbf{G}_{t7}$ by sampling from the prior distribution and decoding, infer a latent transition $\mathbf{z}_{t0}$ with the encoder, or evaluate the likelihood of a given $\mathbf{z}_{t0}$ with the conditional prior. To maintain structural regularity, the output of our decoder is always 20 frames, representing the entire trajectory, but we only use the last 12 frames as the prediction result.}
    \label{fig:CVAE}
\end{figure*}

\textbf{Main Structure} We propose a CVAE which formulates the trajectories $p_{\theta } (\mathbf{G}_{t_n...t_{n+m}}\mid\mathbf{G}_{t_0...t_{n-1}})$ as a latent variable model.
Specifically, similar to the strategy of the previous works \cite{alahi2016social,mohamed2020social}, the prediction module, i.e. the decoder, observes the trajectory of 3.2 seconds which corresponds to 8 frames and predicts the trajectories for the next 4.8 seconds that are 12 frames. Therefore, we use a total of 20 frames of data during training, the CVAE structure is shown in Figure.\ref{fig:CVAE}. 
Following the original CVAE derivation \cite{sohn2015learning}, our model contains two main components. 
First, conditioned on the previous graphs $\mathbf{G}_{t_0...t_7}$, the distribution over possible latent variables $\mathbf{z}_{t_0}$ is described by a learned conditional prior, as shown in Equation \ref{Eq:prior}:
\begin{equation}
    p_{\theta } (\mathbf{z}_{t_0}\mid\mathbf{G}_{t_0...t_7}) = \mathcal{N}(\mathbf{z}_{t_0};\mu_\theta(\mathbf{G}_{t_0...t_7}),\sigma_\theta(\mathbf{G}_{t_0...t_7}))
    \label{Eq:prior}
\end{equation}
which parameterize a Gaussian distribution with diagonal covariance. Intuitively, the latent variable $\mathbf{z}_{t_0}$ characterizes the transition to $\mathbf{G}_{t_8...t_{19}}$ and, as a result, exhibits distinct distributions depending on the value of $\mathbf{G}_{t_0...t_7}$. Incorporating a learned conditional prior enhances the CVAE's capacity to generalize across diverse trajectories and contributes to the stability of training.
Second, given $\mathbf{z}_{t_0}$ and $\mathbf{G}_{t_0...t_7}$ as conditions, the decoder generates the output $\mathbf{G}_{t_8...t_{19}}$, which is the future graph of the pedestrians.
The complete probability model for a transition is then denoted in Equation \ref{Eq:transition}:
\begin{multline}
    p_{\theta}(\mathbf{G}_{t_8...t_{19}}\mid\mathbf{G}_{t_0...t_7}) = \\
    \int_{\mathbf{z}_{t_0}} p_{\theta }(\mathbf{z}_{t_0}\mid\mathbf{G}_{t_0...t_7})p_{\theta }(\mathbf{G}_{t_8...t_19}\mid\mathbf{z}_{t_0},\mathbf{G}_{t_0...t_7})
    \label{Eq:transition}
\end{multline}
To facilitate training, the encoder, serving as an approximate posterior, is incorporated and parameterized a Gaussian distribution follows in Equation \ref{Eq:posterior}: 
\begin{multline}
    q_{\phi} (\mathbf{z}_{t_0}\mid\mathbf{G}_{t_0...t_7},\mathbf{G}_{t_8...t_{19}}) = \\
    \mathcal{N}(\mathbf{z}_{t_0};\mu_\phi(\mathbf{G}_{t_0...t_7},\mathbf{G}_{t_8...t_{19}}),\sigma_\phi(\mathbf{G}_{t_0...t_7},\mathbf{G}_{t_8...t_{19}}))
    \label{Eq:posterior}
\end{multline}
Our CVAE is trained using pairs of $(\mathbf{G}_{t_0...t_7}, \mathbf{G}_{t_8...t_19})$. We then consider the usual variational lower bound as shown in Equation \ref{Eq:lower_bound}:
\begin{multline}
    \log_{}{}{p_{\theta}(\mathbf{G}_{t_8...t_{19}}\mid\mathbf{G}_{t_0...t_{7}})}\ge
    \mathbb{E}_{q\phi}[\log_{}{}p_{\theta}(\mathbf{G}_{t_8...t_{19}}\mid\mathbf{G}_{t_0...t_{7}})] \\
    -D_{KL}(q_{\phi}(\mathbf{z}_{t_0}\mid\mathbf{G}_{t_0...t_{7}},\mathbf{G}_{t_8...t_{19}}) 
\parallel p_{\theta}(\mathbf{z}_{t_0}\mid\mathbf{G}_{t_0...t_{7}}))
    \label{Eq:lower_bound}
\end{multline}
Hence, our objective is to find the values for parameters $(\theta, \phi) $that minimize the combined loss function, 
\begin{equation}
    L = L_{rec} + w_{KL} L_{KL}
    \label{Eq:loss}
\end{equation}
across all pairs in our training dataset. Where the $L_{rec}$ represent the reconstruction loss and $L_{KL}$ is the KL divergence. $w_{KL}$ is a weight for cost annealing \cite{bowman2015generating}, which is used to help decrease the posterior collapse or KL divergence vanish.
The reconstruction loss is derived from the negative log-likelihood of a bivariate Gaussian distribution, where the decoder predicts the mean $(\mu_x,\mu_y)$, standard deviations $(\sigma_x,\sigma_y)$, and correlation coefficient $\rho$ for each pedestrian. The loss encourages the model to maximize the likelihood of the ground truth trajectories under the predicted distribution, which is computed as:

\begin{equation}
L_{\text{rec}} = -\frac{1}{N} \sum_{i=1}^N \log \mathcal{N}\left( 
  \mathbf{v}_i^{\text{trgt}} \mid 
  \bm{\mu}_i^{\text{pred}}, 
  \bm{\Sigma}_i^{\text{pred}}
\right)
\end{equation}

\noindent where: 
\begin{itemize} 
\item $\mathbf{v}_i^{\text{trgt}} = ( x_i^{\text{trgt}} \ y_i^{\text{trgt}}  )$
denotes the ground truth pedestrian trajectory.
\item $\bm{\mu}_i^{\text{pred}} = (\mu_{x,i}^{\text{pred}} \ \mu_{y,i}^{\text{pred}} )$
represents the predicted means of the trajectory.
\item $\bm{\Sigma}_i^{\text{pred}}$ is the predicted covariance matrix, defined as:
$\bm{\Sigma}_i^{\text{pred}} = \begin{bmatrix} 
    \sigma_{x,i}^2 & \rho_i \sigma_{x,i} \sigma_{y,i} \\ 
    \rho_i \sigma_{x,i} \sigma_{y,i} & \sigma_{y,i}^2 
  \end{bmatrix} $\end{itemize}

\begin{figure}[h]
    \centering
    \includegraphics[width=0.45\textwidth]{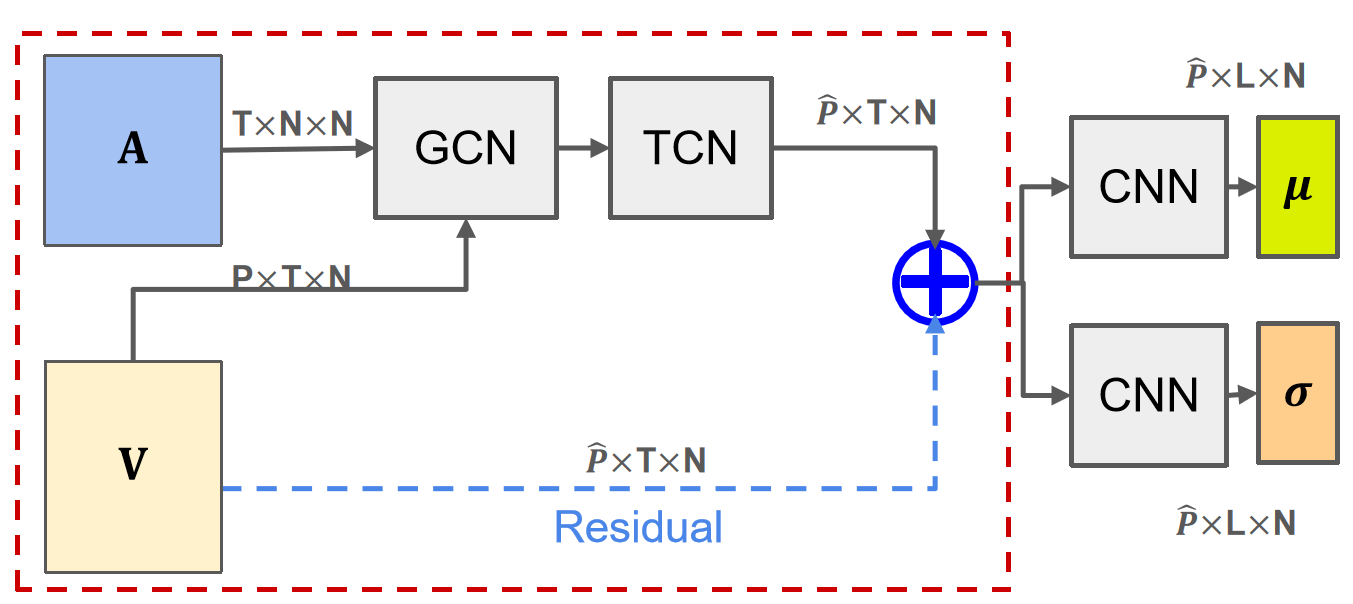}
    \caption{Encoder architecture. The part enclosed by the red dotted line is responsible for extracting the embedding of the spatio-temporal graph. The graph representation is G = (V, A), where $\mathbf{V}$ is set of vertices of the graph, A is the weighted adjacency matrix. $\mathbf{P}$  is the dimension of pedestrian position, $\mathbf{N}$  is the number of pedestrians, $\mathbf{T}$ is the number of time steps and $\mathbf{\hat{P}}$ is the dimensions of the embedding coming from STGCNN, $\mathbf{L}$ is the length of the latent transition.}
    \label{fig:encoder}
\end{figure}

\noindent \textbf{Encoder} The architecture of the encoder is illustrated in Figure. \ref{fig:encoder}. 
The GCNs are used to perform spatial convolutions on graph-structured data, modeling the interactions between entities at each time step. TCNs are then applied to process these spatial graphs across temporal sequences, learning dynamic patterns over time. The residual connection helps superimpose a global information in the embedding. 

In other research based on GNNs, both the input and output are graphs, meaning that the entire architecture typically consists only of convolutional layers, rendering the size of spatial graphs irrelevant. This is crucial because the number of pedestrians in the observed area may vary.
However, a CVAE architecture typically assumes a fixed-dimensional latent representation, often in the form of a one-dimensional vector. This assumption becomes problematic when dealing with variable-sized inputs.
To address this limitation, we propose two parallel convolutional layers that encode the latent mean and variance with flexible spatial dimensions. These layers operate directly on the spatio-temporal graph embeddings without flattening or pooling, thereby preserving the underlying spatial structure. As a result, the model can naturally accommodate variable-sized inputs during training, since the fully convolutional design ensures consistency between input and output dimensions.
During inference, the decoder requires both past graph frames and samples from the prior network. To ensure compatibility, we extract the required dimensionality from the prior network output corresponding to the current temporal segment, enabling coherent decoding. Experimental results in the following section demonstrate the effectiveness of this design.
In our implementation, the latent variable $z_{t0}$ is represented as a transition tensor aligned with the spatio-temporal graph embedding, rather than a flattened vector. This formulation preserves the variable number of agents without resorting to pooling or resizing, while remaining fully compatible with the standard CVAE objective.

As both networks serve as encoders, the architecture of the prior network closely mirrors that of the recognition network, with only minor modifications.
To enhance the expressive capacity of the prior network, we increase the number of GCN and TCN layers. In contrast, the recognition network is kept slightly less expressive to reduce the risk of posterior collapse. Additionally, dropout is applied to the recognition network, and small Gaussian perturbations are added to its output during training to further stabilize optimization.
The detailed architectural differences between the prior and recognition networks are summarized in Table \ref{tab:encoder_compare}.

\begin{table}[h]
\centering
\caption{Comparison between prior and recognition networks}
\begin{tabular}{lcc}
\hline
Component & Prior Network & Recognition Network \\
\hline
Input & Past trajectories & Past + future trajectories \\
GCN layers & 3 & 2 \\
TCN layers & 3 & 2 \\
Dropout & No & Yes \\
Gaussian noise & No & Yes \\
Role & Conditional prior & Approximate posterior \\
\hline
\end{tabular}
\label{tab:encoder_compare}
\end{table}

This asymmetric design follows common practices in variational models to balance representation capacity and training stability.

\begin{figure}[h]
    \centering
    \includegraphics[width=0.45\textwidth]{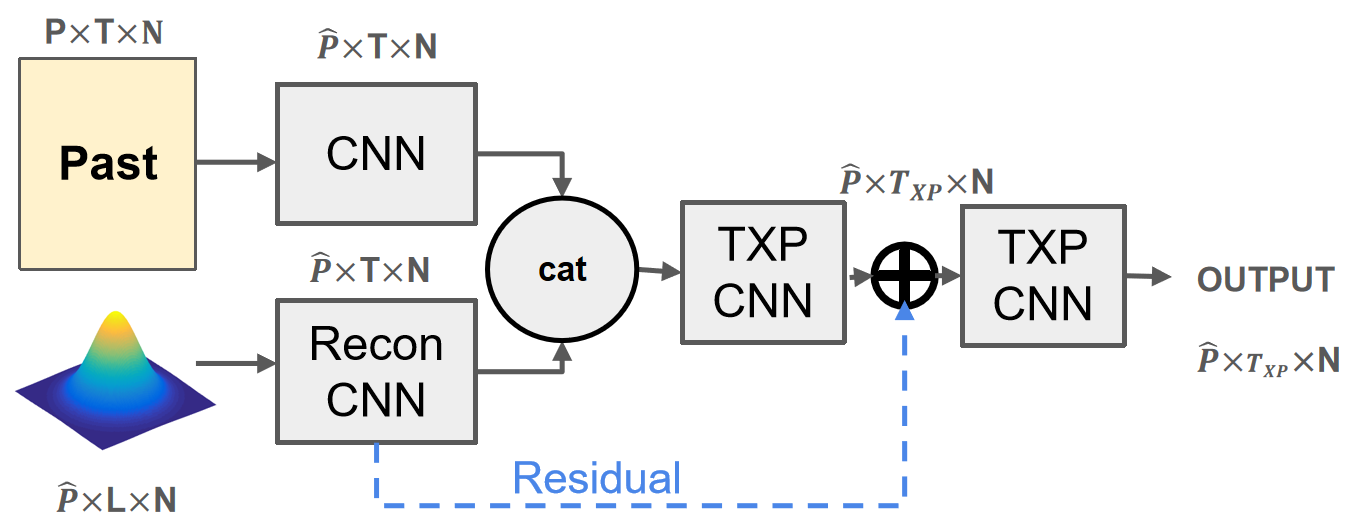}
    \caption{Decoder architecture. The input consists of the past Ground Truth of the past trajectories and the latent transition. TXPCNN is used to extend the time dimension to the expected output dimension. $\mathbf{T_{xp}}$ denotes the output number of time steps.}
    \label{fig:decoder}
\end{figure}

\textbf{Decoder} The architecture of the decoder is illustrated in Figure. \ref{fig:decoder}, 
which consists of two inputs: the past trajectory graphs and the distribution output by the prior network, representing the latent transition. First, two CNN layers are applied to both the ground truth graph and the latent transition, converting them back into graph embeddings. Following the standard approach in CVAE architectures, a cascade method is employed to fuse these two components. This fused embedding is subsequently passed through two Time-Extrapolator Convolutional Neural Networks (TXP-CNNs), which are specifically designed to extrapolate the embedding to the desired output time series length. To emphasize the importance of the latent transition and help prevent posterior collapse, the reconstructed latent transition is skip-connected between the two TXP-CNN blocks. 

\section{Experimental Setup}\label{sec:4}

\subsection{Training Datasets and Evaluation Metrics}

The training of the model involves two human trajectory prediction datasets: ETH \cite{pellegrini2009you} and UCY \cite{lerner2007crowds}. The ETH dataset comprises two scenes labeled as ETH and HOTEL, while the UCY dataset encompasses three scenes identified as ZARA1, ZARA2, and UNIV. The frames within these datasets are sampled at an observation frequency of 2.5 Hz.
As mentioned earlier, when being evaluated, the decoder uses the trajectory of 3.2 seconds which corresponds to 8 frames as the prior and generate the trajectories for the next 4.8 seconds with 12 frames in total.
Two metrics are employed to assess the model's performance: the Final Displacement Error (FDE), as introduced in \cite{alahi2016social} and the Average Displacement Error (ADE), as introduced in \cite{pellegrini2009you}. In essence, FDE evaluates the precision of predictions specifically at the endpoint, whereas ADE gauges the average prediction accuracy across the trajectory.

The model ultimately produces a bi-variate Gaussian distribution as its prediction. 
To evaluate the predicted distributions, 20 samples are drawn from the predicted distribution and the sample closest to the ground truth is used to compute ADE and FDE.
We follow the widely-used ETH/UCY protocol adopted by Social-LSTM/GAN/STGCNN for fair comparison. We acknowledge that best-of-K evaluation favors multimodal predictors; complementary likelihood or calibration metrics are valuable future work, while this paper focuses on endpoint robustness and real-time deployability under the established benchmark protocol.
%
This evaluation method has been widely used in similar tasks and has been previously applied in various works, including Social-LSTM, Social-GAN, Social-STGCNN, etc.

\subsection{Model configuration and training setup}
We set a training batch size of 128 and trained the model for 250 epochs using Stochastic Gradient Descent (SGD). The initial learning rate is 0.01, and changed to 0.002 after 150 epochs. The weight $w_{KL}$ follows a linear annealing schedule from 0 to 2 × $10^{-5}$ × epochs. This gradual increase prevents the KL term from dominating the loss in early training stages, allowing the model to first focus on reconstruction before regularizing the latent space—a strategy proven effective in avoiding posterior collapse \cite{bowman2015generating}.
In essence, if the length of the latent transition is increased, more features of the graph can be captured, leading to greater separation between the numbers and potentially improving recognition performance. However, as indicated in \cite{doersch2016tutorial}, the latent transition length performs effectively within the range of 4 to 1000. In other words, VAE/CVAE exhibits a robustness to changes in the latent variable dimension.
In our specific case, after trial and error, we defined the latent transition length to 20 as it yielded the best results. It is worth mentioning that while the ``length" here is fixed, other dimensions of the latent transition are not fixed and vary with the number of pedestrians.
All experiments were conducted on A100 GPU in Google Colab.

\subsection{Robot-Collected Dataset Setup}

\begin{figure*}[t]
  \centering
  \begin{subfigure}{0.30\textwidth}
    \centering
    \includegraphics[width=\linewidth]{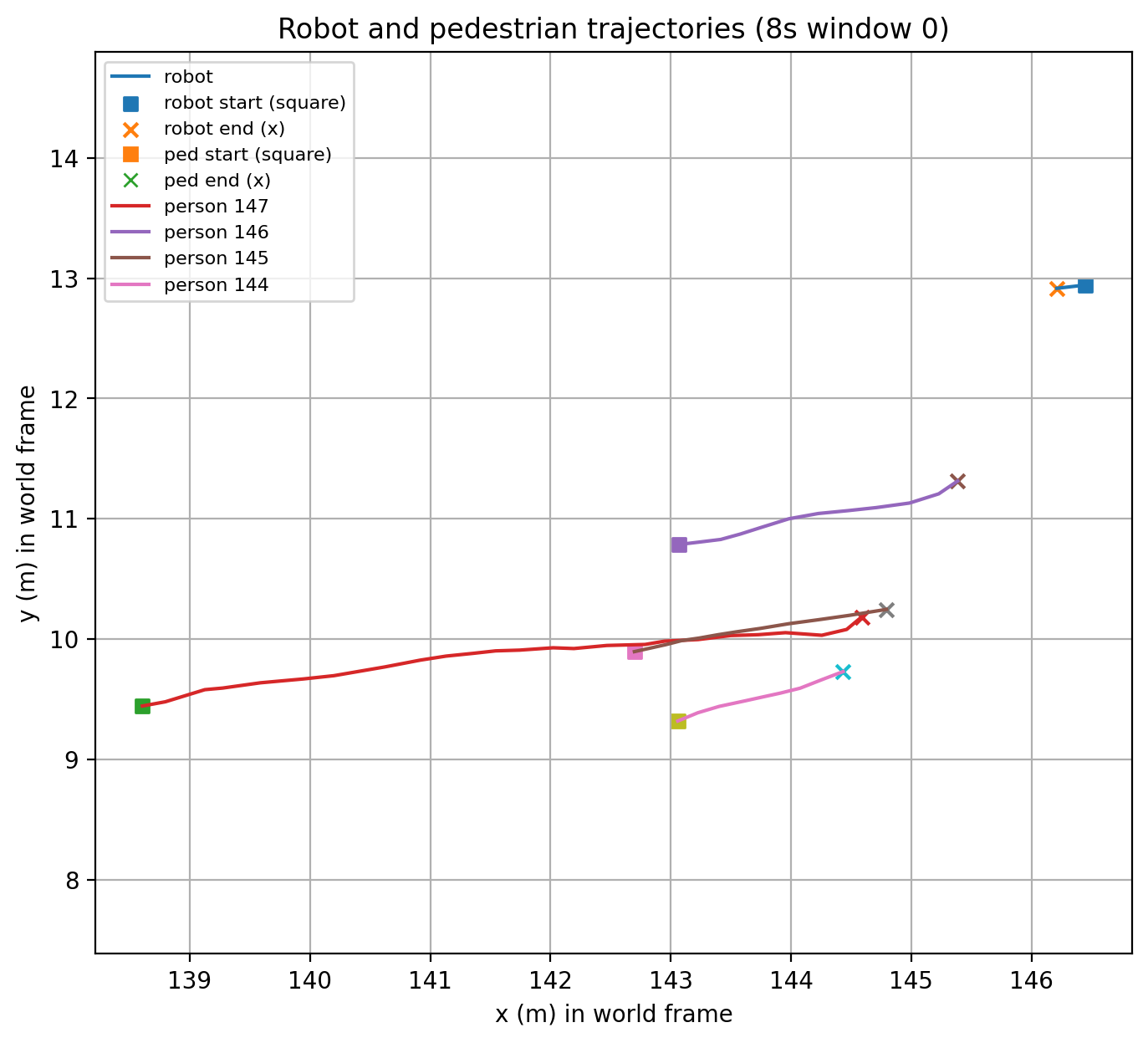}
  \end{subfigure}\hfill
  \begin{subfigure}{0.30\textwidth}
    \centering
    \includegraphics[width=\linewidth]{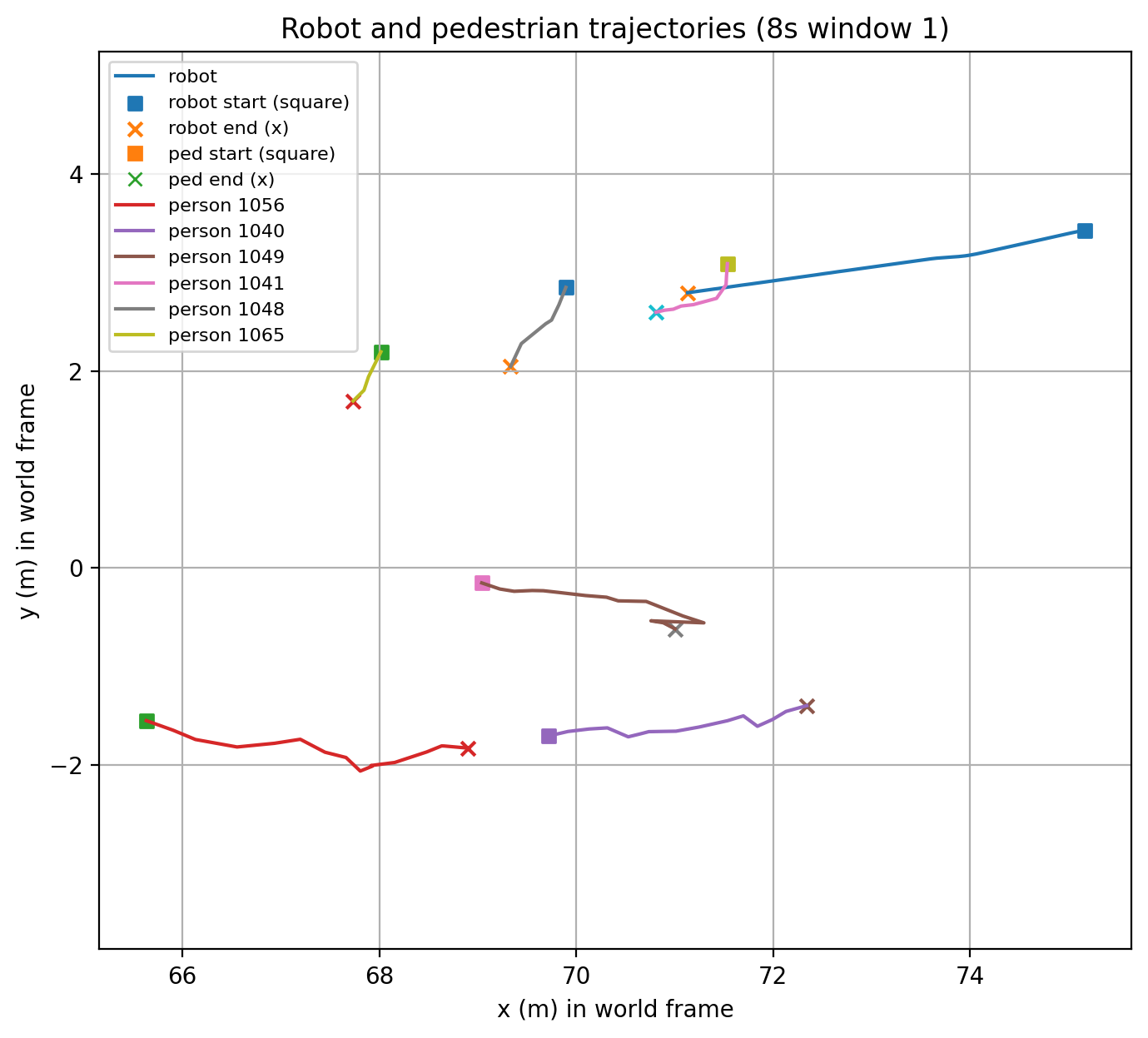}
  \end{subfigure}\hfill
  \begin{subfigure}{0.30\textwidth}
    \centering
    \includegraphics[width=\linewidth]{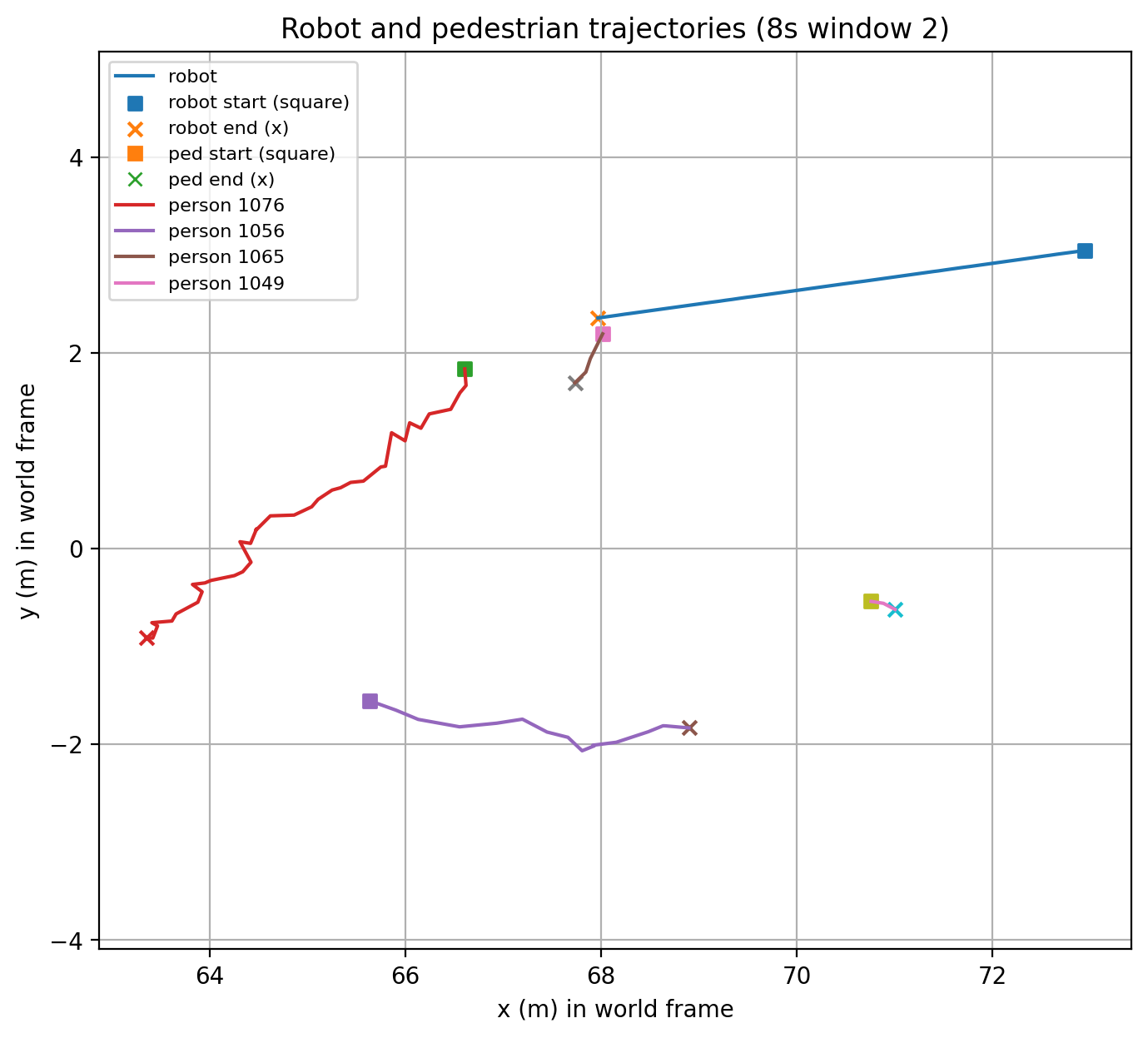}
  \end{subfigure}
  \caption{Examples of robot and pedestrian trajectories over multiple 8\,s windows. Squares denote starting points and crosses denote endpoints. The plots illustrate typical characteristics of the collected data, including short and partially observed trajectories, varying motion patterns, and interactions around a moving robot.}
  \label{fig:dataset_example}
\end{figure*}

To evaluate the model under more realistic conditions, we additionally conduct experiments on a pedestrian trajectory dataset \cite{certad2025v2p} collected using a mobile robotic platform. The data is recorded in a semi-open campus area of Johannes Kepler University Linz.
Following the standard ETH/UCY protocol, the dataset is segmented into short trajectory sequences of 20 frames, consisting of 8 observed frames and 12 prediction frames (corresponding to 3.2 seconds of observation and 4.8 seconds of prediction). To ensure temporal consistency with this setup, the raw trajectories are resampled to a fixed rate of 2.5 Hz during preprocessing, such that each 20-frame sequence spans a total of 8 seconds. Pedestrian trajectories are transformed from robot-centric observations to the world frame using odometry. 
Figure~\ref{fig:dataset_example} illustrates several representative examples from the dataset. As shown in this figure, we also incorporate the robot’s own trajectory as part of the trajectory graph input. This allows the model to capture potential interaction cues between pedestrians and the robot, which can contribute to more accurate trajectory prediction.

\section{Results and Discussion}\label{sec:5}

\begin{table*}[h]
\centering
\resizebox{\textwidth}{!}{
\begin{tabular}{l||l|l|l|l|l|l}
\hline
                          & ETH       & HOTEL     & ZARA1    & ZARA2    & UNIV      & AVG       \\ \hline
\hline
S-LSTM \cite{alahi2016social}                    & 1.09 / 2.35 & 0.79 / 1.76 & 0.47 / 1.00 & 0.56 / 1.17 & 0.67 / 1.40 & 0.72 / 1.54 \\ \hline
S-GAN-P \cite{gupta2018social}                   & 0.87 / 1.62 & 0.67 / 1.37 & 0.35 / 0.68 & 0.42 / 0.84 & 0.76 / 1.52 & 0.61 / 1.21 \\ \hline
Sophie \cite{sadeghian2019sophie}                & 0.70 / 1.43 & 0.76 / 1.67 & \textbf{0.30} / 0.63 & 0.38 / 0.78 & 0.54 / 1.24 & 0.54 / 1.15 \\ \hline
CGNS \cite{li2019conditional}                      & \textbf{0.62} / 1.40 & 0.70 / 0.93 & 0.32 / 0.59 & 0.35 / 0.71 & 0.48 / 1.22 & 0.49 / 0.97 \\ \hline
Social-BiGAT \cite{kosaraju2019social}             & 0.69 / 1.29 & 0.49 / 1.01 & \textbf{0.30} / 0.62 & 0.36 / 0.75 & 0.55 / 1.32 & 0.48 / 1.00 \\ \hline
AST-GNN \cite{zhou2021ast}                  & 0.66 / 1.02 & \textbf{0.37} / 0.61 & 0.32 / 0.52 & \textbf{0.28} / 0.45 & 0.46 / 0.83 & \textbf{0.42} / 0.69 \\ \hline
\hline
Social-STGCNN(Paper)      & 0.64 / 1.11 & 0.49 / 0.85 & 0.34 / 0.53 & 0.30 / 0.48 & \textbf{0.44} / 0.79 & 0.44 / 0.75 \\ \hline
Social-STGCNN(Experiment) & 0.74 / 1.23 & 0.41 / 0.68 & 0.33 / 0.52 & 0.30 / 0.48 & 0.49 / 0.91 & 0.45 / 0.76 \\ \hline
Our CVAE                  & 0.73 / \textbf{0.93} & 0.40 / \textbf{0.55} & 0.34 / \textbf{0.44} & 0.31 / \textbf{0.41} & 0.47 / \textbf{0.69} & 0.45 / \textbf{0.60} \\ \hline
\end{tabular}}
\caption{The ADE/FDE metrics for various methods compared to Social-STGCNN-CVAE are presented, including both the original Social-STGCNN paper's reported results and our experimentally obtained results. Notably, when using the experimental results of Social-STGCNN as the baseline, our model lags even less in ADE.}
\label{tab:compare}
\end{table*}

\subsection{Inference speed and model size}
The number of trainable parameters in the model is significantly influenced by the latent transition length, as mean and variance are encoded using two separate CNNs. We examined various latent transition lengths and evaluated their performance using the ETH dataset. Table \ref{tab:z_size} provides a brief comparison of how different latent transition lengths affect the number of parameters and inference time, compared to the baseline model Social-STGCNN, as well as their impact on performance.
The latent transition with length of 20 achieves the optimal balance between accuracy and efficiency (0.73 ADE/0.93 FDE). Experiments with varying latent space lengths (e.g., 10 or 30) show worse performance (0.79/1.10 and 0.86/1.07, respectively), indicating that a shorter or longer latent transition length leads to a decline in performance. The best model achieves an inference time of 0.0022 seconds per step. Although this is twice as slow as Social-STGCNN, it remains efficient enough for real-time applications. The significant improvement in FDE justifies this trade-off, especially in safety-critical scenarios. 

\begin{table}[]
\begin{tabular}{|l|l|l|l|}
\hline
              & Parameters count & Inference time (s) & ADE/FDE   \\ \hline
Social-STGCNN  & 7.6K                & 0.0011        & \textbf{0.64}/1.11 \\ \hline
Ours z\_len=10 & 18.5K               & 0.0022        & 0.79/1.10 \\ \hline
Ours z\_len=20 & 24.6k               & 0.0024        & 0.73/\textbf{0.93}  \\ \hline
Ours z\_len=30 & 30.8K               & 0.0025        & 0.86/1.07  \\ \hline
\end{tabular}
\caption{Compared to the original Social-STGCNN, the impact of our model with different lengths of latent space on the number of parameters, inference time, and accuracy on ETH dataset.}
\label{tab:z_size}
\end{table}

It is also worth mentioning that alternative strategies, such as applying a Spatial Pyramid Pooling (SPP) layer \cite{he2015spatial} after the embedding extraction to convert spatial graphs of varying sizes into a uniform representation, followed by a fully connected layer to project them into a one-dimensional latent code, as is commonly done in CVAE architectures, were explored during the early stages of model development but were ultimately abandoned. This is because applying aggressive pooling on spatial graphs with only a few pedestrian nodes results in significant information loss, making it difficult for the latent space to effectively capture the underlying probabilistic distribution. Consequently, the decoder tends to ignore the latent variables altogether, causing the latent space to become uninformative. Since decoding from a standard normal distribution yields outputs nearly indistinguishable from those generated using the prior, the latent representation becomes effectively redundant. This limitation highlights the practical challenges of applying standard CVAE techniques to dynamic spatial graphs, and underscores the strength of our approach in maintaining informative latent representations in real-world pedestrian scenarios. 

\subsection{Comparative Evaluation}
The performance of our model is compared with various methods as well as the original Social-STGCNN and on ADE/FDE metrics in Table \ref{tab:compare}. 
As can be seen, although each model demonstrates its own strengths, our model still achieves some of the highest scores. Since our primary hypothesis centers on the idea that integrating the CVAE architecture can enhance the performance of the original Social-STGCNN, we place particular emphasis on comparing our results with those of the Social-STGCNN baseline. 
We observed discrepancies between the performance reported in the original Social-STGCNN paper and the results obtained using the publicly available implementation and pretrained models, a reproducibility gap that has also been noted in prior works. To ensure fairness and consistency, we conducted multiple runs across different environments and obtained stable results that differ from the originally reported numbers. Therefore, we report both the original results and those reproduced under our experimental setup, using the latter as the primary baseline for comparison. Although these reproduced results are slightly lower than the original values, they provide a more reliable and consistent reference. Presenting both sets of results enhances transparency and enables a fair comparison under unified evaluation protocols.

Compared to Social-STGCNN, our model achieves competitive performance across all scenarios. Notably, in the ETH scene, despite a marginal increase in ADE (ours 0.73 vs. 0.64/0.74), the FDE is significantly reduced by 16-24\% (ours 0.93 vs. 1.11/1.23). Given that the authors of the original Social-STGCNN paper noted that merely increasing the number of network layers does not lead to performance improvements, our results suggest that the introduction of a CVAE framework suggests improved modeling of longer-term motion patterns. This, in turn, contributes to more accurate endpoint predictions. Similar trends are observed in the HOTEL and UNIV scenes, where FDE improves by around 19\% and 24\%, respectively. These results demonstrate that the CVAE framework demonstrates the benefits of incorporating probabilistic modeling, where the conditional prior distribution enables robust trajectory predictions by explicitly encoding uncertainties in pedestrian motion and social interactions.
While ADE captures average tracking accuracy, in navigation and collision checking the terminal position over a 4.8s horizon is often the dominant factor. Our CVAE formulation primarily reduces FDE (e.g., 0.60 vs 0.76 average on ETH/UCY) with comparable ADE, indicating improved endpoint robustness without sacrificing overall trajectory fidelity.

\begin{figure}[h]
    \centering
    \includegraphics[width=0.45\textwidth]{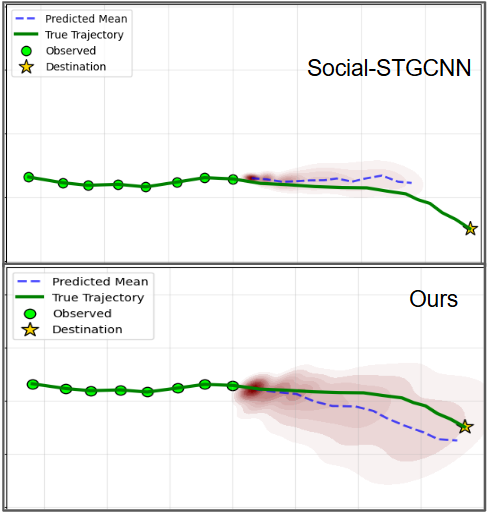}
    \caption{A representative example of trajectory distribution prediction: the top shows Social-STGCNN, while the bottom shows our method.}
    \label{fig:qualitative}
\end{figure}

Figure \ref{fig:qualitative} shows a qualitative comparison of our model and Social-STGCNN. 
As shown in Figure \ref{fig:qualitative}, the predicted distribution of our model is wider, indicating a higher degree of uncertainty awareness compared to the deterministic baseline. At the same time, the probability mass is better aligned with the true endpoint, resulting in improved FDE. This behavior suggests that the conditional prior in the CVAE effectively captures multiple plausible future motions while maintaining accurate endpoint prediction, which is particularly important for safety-critical planning scenarios.

\subsection{Validation on Robot-Collected Dataset}

To further evaluate the model under realistic conditions, we conduct additional validation on a pedestrian trajectory dataset collected using a mobile robotic platform. Compared to benchmark datasets such as ETH/UCY, this dataset presents additional challenges, including shorter trajectory lengths, occlusions, sensor noise, and irregular sampling.
We directly evaluate the pretrained models on this dataset without additional fine-tuning or domain-specific adaptation. This setup allows us to assess the generalization capability of the models under real-world sensing conditions.
Table \ref{tab:robot} reports the performance of Social-STGCNN and our method on the robot-collected dataset. These results are obtained by directly applying the pretrained models without any additional training or adaptation on this dataset.
As can be observed, both models experience a performance degradation compared to benchmark datasets, which is expected due to the increased difficulty of real-world data. However, our method maintains more stable endpoint predictions and achieves lower FDE compared to the baseline.

\begin{table}[h]
\centering
\begin{tabular}{|l|c|c|}
\hline
Model & ADE & FDE \\
\hline
Social-STGCNN & 0.98& 1.55 \\
Ours (CVAE)   & 1.01& 1.29 \\
\hline
\end{tabular}
\caption{Performance comparison on the robot-collected dataset.}
\label{tab:robot}
\end{table}

These results suggest that incorporating probabilistic modeling helps improve robustness to noise and uncertainty in real-world scenarios. In particular, the CVAE formulation enables the model to better handle ambiguous or partially observed trajectories. This behavior is particularly relevant for real-world applications such as modular buses and mobile robots operating in dynamic environments. 

\section{Conclusion and Future Work}\label{sec:6}

In this paper, we presented a CVAE-based extension of the Social-STGCNN framework for multimodal pedestrian trajectory prediction. By integrating a probabilistic formulation into a graph-based convolutional architecture, the proposed approach is able to generate diverse future trajectories while maintaining computational efficiency.

We evaluated the method on the ETH/UCY benchmark datasets as well as on a real-world pedestrian dataset collected using a mobile robotic platform. Experimental results show moderate improvements in FDE compared to the Social-STGCNN baseline, while maintaining comparable ADE. The model demonstrates more consistent endpoint predictions and improved trajectory diversity across different crowd configurations. Despite a slight increase in inference time, the method remains suitable for real-time applications.
We further demonstrate the effectiveness of the proposed model under real-world conditions through evaluation on a robot-collected pedestrian dataset.

For future work, several directions can be explored. First, incorporating additional contextual information, such as scene semantics or environmental constraints, may further improve prediction accuracy. Second, integrating higher-level behavioral cues, such as pedestrian intent or group interactions, could help better capture complex motion patterns. Finally, extending the framework to explicitly model interactions between pedestrians and robotic platforms may improve its applicability in tightly coupled human-robot environments.

\section*{ACKNOWLEDGMENT}
This work was supported by the Ergodic project, co-financed by the European Union and Austrian Research Promotion
Agency (FFG), project number: 905494

\bibliographystyle{IEEEtran}
\bibliography{bib}
\end{document}